\documentclass[11pt,letterpaper]{article}
\usepackage{emnlp2016}
\usepackage{times}
\usepackage{latexsym}
\usepackage{amsfonts}
\usepackage{amsmath}
\usepackage{graphics}
\usepackage{amssymb}
\usepackage{url}
\usepackage{multirow}
\usepackage{subfig}
\usepackage[export]{adjustbox}
\usepackage{algorithmicx}
\usepackage[noend]{algpseudocode}
\usepackage{amsthm}

\usepackage[first=0,last=9]{lcg}

\usepackage{url,color,mdframed}

\usepackage[table]{xcolor}

\usepackage{lipsum}

\emnlpfinalcopy

\def\emnlppaperid{***}

\newtheorem{mydef}{Definition}
\newcommand{\gralap}{\texttt{GraLap}}

\title{All Fingers are not Equal: Intensity of References in Scientific Articles\\({\color{blue}Accepted in EMNLP, 2016})}

\author{Tanmoy Chakraborty \\ Dept. of Computer Science \& UMIACS\\  University of Maryland, College Park, USA\\ tanchak@umiacs.umd.edu 
        \And 
Ramasuri Narayanam \\ IBM Research, India \\ ramasurn@in.ibm.com}

\date{}

\begin{document}

\maketitle

\begin{abstract}
Research accomplishment is usually measured by considering all citations with equal importance, thus ignoring the wide variety of purposes an article is being cited for. Here, we posit that measuring the intensity of a reference is crucial not only to perceive better understanding of research endeavor, but also to improve the quality of citation-based applications. To this end, we collect a rich annotated dataset with references labeled by the intensity, and propose a novel graph-based semi-supervised model, \gralap~ to label the intensity of references. Experiments with AAN datasets show a significant improvement compared to the baselines to achieve the true labels of the references (46\% better correlation). Finally, we provide four applications to demonstrate how the knowledge of reference intensity leads to design better real-world applications.
\end{abstract}

\section{Introduction}
With more than one hundred thousand new scholarly articles being published each year, there is a rapid growth in the number of citations for the relevant
scientific articles. In this context, we highlight the following interesting facts about the process of citing scientific articles: (i) the most commonly cited paper by Gerard Salton, titled  ``A Vector Space Model for Information Retrieval'' (alleged to have been published in 1975) does not actually exist in reality \cite{Dubin04}, (ii) the scientific authors read only 20\% of the works they cite \cite{Simkin_Roychowdhury:2003}, (iii) one third of the references in a paper are redundant and 40\% are perfunctory \cite{Moravcsik}, (iv) 62.7\% of the references could not be attributed a specific function (definition, tool etc.) \cite{Teufel:2006}. Despite these facts, the existing bibliographic metrics consider that all citations are {\em equally significant}.

In this paper, we would emphasize the fact that {\em all the references of a paper are not equally influential}. For instance, we believe that for our current paper, \cite{WanL14} is more influential reference than \cite{Eugene}, although the former has received lower citations (9) than the latter (1650) so far\footnote{The statistics are taken from Google Scholar on June 2, 2016.}. Therefore the influence of a cited paper completely depends upon the context of the citing paper, not the overall citation count of the cited paper.  We further took the opinion of the original authors of few selective papers and realized that around 16\% of the references in a paper are highly influential, and the rest are trivial (Section \ref{data}). This motivates us to design a prediction model, \gralap~ to automatically label the influence of a cited paper with respect to a citing paper. Here, we label paper-reference pairs rather than references alone, because a reference that is influential for one citing paper may not be influential with equal extent for another citing paper. 

We experiment with ACL Anthology Network (AAN) dataset and show that \gralap~ along with the novel feature set, quite efficiently, predicts the intensity of references of papers, which achieves (Pearson)  correlation of $0.90$  with the human annotations.
Finally, we present four interesting applications to show the efficacy of considering unequal intensity of references, compared to the uniform intensity.

The contributions of the paper are four-fold: (i) we acquire a rich annotated dataset where paper-reference pairs are labeled based on the influence scores (Section \ref{data}), which is perhaps the first gold-standard for this kind of task; (ii) we propose a graph-based label propagation model \gralap~for semi-supervised learning which has tremendous potential for any task where the training set is less in number and labels are non-uniformly distributed (Section \ref{model}); (iii) we propose a diverse set of features (Section \ref{feature}); most of them turn out to be quite effective to fit into the prediction model and yield improved results  (Section \ref{sec:result}); (iv)  we present four applications to show how incorporating the reference intensity enhances the performance of several state-of-the-art systems (Section \ref{appli}).

\section{Defining Intensity of References} \label{intensity}

All the references of a paper usually do not carry equal intensity/strength with respect to the citing paper  because some papers
have influenced the  research more than others. 
To pin down this intuition, 
here we discretize the reference intensity by numerical values within the range of $1$ to $5$, ($5$: most influential, $1$: least influential). The appropriate definitions of different labels of reference intensity are presented in Figure \ref{intensity_def}, which are also the basis of building the annotated dataset (see Section \ref{data}):

\begin{figure}[!htb]
\caption{Definitions of the intensity of references}\label{intensity_def}
  \begin{mdframed}[backgroundcolor=gray!15]
{\small
 $\bullet$ {\bf Label-1:} The reference is related to the citing article with {\em very limited extent} and can be {\em removed} without compromising the competence of the references (e.g., \cite{Eugene} for this paper).\\
 $\bullet$ {\bf Label-2:} The reference is {\em little mentioned} in the citing article and can be {\em replaced} by others without compromising the adequacy of the references (e.g., \cite{ASI:ASI23179} for this paper).\\
 $\bullet$ {\bf Label-3:} The reference occurs separately in a sentence within the citing article and has {\em no significant impact on the current problem} (e.g., references to metrics, tools) (e.g., \cite{Porter:1997} for this paper).\\
  $\bullet$ {\bf Label-4:} The reference is {\em important} and highly related to the citing article. It is usually
mentioned several times in the article with long reference context (e.g., \cite{Singh:2015} for this paper).\\
$\bullet$ {\bf Label-5:} The reference is {\em extremely important} and occurs (is emphasized) multiple times within the citing article. It generally points to the cited article from where the citing article borrows main ideas (and can be treated as a baseline) (e.g., \cite{WanL14} for this paper).}
\end{mdframed}
\end{figure}

 Note that ``reference intensity'' and ``reference similarity'' are two different aspects. It might happen that two similar reference  are used with different intensity levels in a citing paper -- while one is just mentioned somewhere in the paper and other is used as a baseline. Here, we address the former problem as a semi-supervised learning problem with clues taken from content of the citing and cited papers.

 \section{Reference Intensity Prediction Model}\label{model}
 In this section, we formally define the problem and introduce our prediction model.
 
 \subsection{Problem Definition}
 We are given a set of papers $\mathbb{P}=\{P_1,P_2,...,P_M\}$ and a sets of references $\mathbb{R}=\{R_1,R_2,...,R_M\}$, where $R_i$ corresponds to the set of references (or cited papers) of $P_i$. There is a set of papers $P_L \in \mathbb{P}$ whose references $R_L\in \mathbb{R}$ are already labeled by $\ell \in L=\{1,...,5\}$ (each reference is labeled with exactly one value).
Our objective is to define a predictive function $f$ that labels the references $R_U\in \{\mathbb{R} \setminus R_L\}$ of the papers $P_U\in \{\mathbb{P} \setminus P_L\}$ whose reference intensities are unknown, i.e., $f: (\mathbb{P}, \mathbb{R}, P_L, R_L, P_U, R_L) \longrightarrow L$.

Since the size of the annotated (labeled) data is much smaller than unlabeled data ($|P_L|\ll|P_U|$), we consider it as a semi-supervised learning problem.

\begin{mydef}
 {\bf (Semi-supervised Learning)} Given a set of entries $X$ and a set of possible labels $Y_L$, let us assume that ($x_1,y_1$), ($x_2,y_2$),..., ($x_l,y_l$) be the set of labeled data where $x_i$ is a data point and $y_i \in Y_L$ is its corresponding label. We assume that at least one instance of each class label is present in the labeled dataset. Let ($x_{l+1},y_{l+1}$), ($x_{l+2},y_{l+2}$),..., ($x_{l+n},y_{l+u}$) be the unlabeled data points where $Y_U={\{y_{l+1},y_{l+2},...y_{l+u}\}}$ are unknown. Each entry $x\in X$ is represented by a set of features $\{f_1,f_2,...,f_D\}$. The problem is to determine the unknown labels using $X$ and $Y_L$.
\end{mydef}


\subsection{\gralap: A Prediction Model}
We propose \gralap, a variant of label propagation (LP) model proposed by \cite{Zhu03semi} where a node in the graph propagates its associated label to its neighbors based on the proximity. We intend to assign same label to the vertices which are closely connected. 
However unlike the traditional LP model where  the original values of the labels continue to fade as the algorithm progresses, we systematically handle this problem in \gralap. Additionally, we follow a post-processing in order to handle ``class-imbalance problem''.\\
\noindent{\bf Graph Creation.} The algorithm starts with the creation of a {\em fully connected weighted graph $G=(X,E)$} where nodes are data points and the weight $w_{ij}$  of each edge $e_{ij}\in E$ is determined by the  radial basis function  as follows:
\vspace{-2mm}
\begin{equation}\small
 w_{ij}=exp\bigg( - \frac{\sum_{d=1}^D (x_i^d - x_j^d)^2}{\sigma^2}  \bigg)
\end{equation}

The weight is controlled by a parameter $\sigma$. Later in this section, we shall discuss how $\sigma$ is selected.  Each node is allowed to propagate its label to its neighbors through edges (the more the edge weight, the easy to propagate). \\
\noindent{\bf Transition Matrix.} We create a probabilistic transition matrix $T_{|X|\times |X|}$, where each entry $T_{ij}$ indicates the probability of jumping from $j$ to $i$ based on the following: $T_{ij}=P(j\rightarrow i)= \frac{w_{ij}}{\sum_{k=1}^{|X|} w_{kj}}$.   \\
\noindent{\bf Label Matrix.} Here, we allow a soft label (interpreted as a distribution of labels) to be associated with each node. We then define a label matrix $Y_{|X|\times |L|}$, where $i$th row indicates the label distribution for node $x_i$. Initially, $Y$ contains only the values of the labeled data; others are zero.\\
\noindent{\bf Label Propagation Algorithm.} This algorithm works as follows:

\begin{figure}[htb]
  \begin{mdframed}[backgroundcolor=gray!10]
    \begin{algorithmic}[1] \small

      \State Initialize $T$ and $Y$
      \While{($Y$ does not converge)}
          \State $Y \leftarrow TY$
          \State Normalize rows of $Y$, $y_{ij}=\frac{y_{ij}}{\sum_k y_{ik}}$
          \State Reassign original labels to $X_L$
      \EndWhile
    \end{algorithmic}
  \end{mdframed}
\vspace{-0.5cm}\end{figure}

After initializing $Y$ and $T$, the algorithm starts by disseminating the label from one node to its neighbors (including self-loop) in one step (Step 3). Then we normalize each entry of $Y$ by the sum of its corresponding row in order to maintain the interpretation of label probability (Step 4). Step 5 is crucial; here we want the labeled sources $X_L$ to be persistent. During the iterations, the initial labeled nodes $X_L$ may fade away with other labels. Therefore we forcefully restore their actual label by setting $y_{il}=1$ (if $x_i\in X_L$ is originally labeled as $l$), and other entries ($\forall_{j\neq l}y_{ij}$) by zero. We keep on ``pushing'' the labels from the labeled data points which in turn pushes the class boundary through high density data points and settles in low density space.  In this way, our approach intelligently uses the unlabeled data in the intermediate steps of the learning. \\
\noindent{\bf Assigning Final Labels.} Once $Y_U$ is computed, one may take the most likely label from the label distribution for each unlabeled data. However, this approach does not guarantee the label proportion observed in the annotated data (which in this case is not well-separated as shown in Section \ref{data}). Therefore, we adopt a {\em label-based normalization} technique. Assume that the label proportions in the labeled data are $c_1,...,c_{|L|}$ (s.t. $\sum_{i=1}^{|L|} c_i=1)$. In case of $Y_U$, we try to balance the label proportion observed in the ground-truth. The label mass is the column sum of $Y_U$, denoted by $Y_{U_{.1}},...,Y_{U_{.|L|}}$, each of which is scaled in such a way that $Y_{U_{.1}}:...:Y_{U_{.|L|}}=c_1:...:c_{|L|}$. The label of an unlabeled data point is finalized as the label with maximum value in the row of $Y$.\\
\noindent{\bf Convergence.} Here we briefly show that our algorithm is guaranteed to converge. Let us combine Steps 3 and 4 as $Y \leftarrow \hat{T}Y$,  where $\hat{T}=T_{ij}/{\sum_{k} T_{ik}}$. $Y$ is composed of $Y_{L_{l\times |L|}}$ and $Y_{U_{u\times |L|}}$, where $Y_U$ never changes because of the reassignment. We can split $\hat{T}$ at the boundary of labeled and unlabeled data as follows:
\vspace{-3mm}
\begin{align*}\scriptsize
  \hat{F} &=  \begin{bmatrix}
    \hat{T}_{ll} & \hat{T}_{lu} \\
    \hat{T}_{ul}      & \hat{T}_{uu}
  \end{bmatrix}
\end{align*}

Therefore, $Y_U \leftarrow \hat{T}_{uu}Y_U + \hat{T}_{ul}Y_L$, which can lead to $Y_U=  \lim_{n\to\infty} \hat{T}_{uu}^n Y^0 + [\sum_{i=1}^n  \hat{T}_{uu}^{(i-1)}] \hat{T}_{ul}Y_L$, where $Y^0$ is the shape of $Y$ at iteration $0$. We need to show $\hat{T}_{uu_{ij}}^n Y^0 \leftarrow 0$. By construction, $\hat{T}_{ij}\geq 0$, and since $\hat{T}$ is row-normalized, and $\hat{T}_{uu}$ is a part of $\hat{T}$, it leads to the following condition: $\exists \gamma < 1,\ \sum_{j=1}^u\hat{T}_{uu_{ij}} \leq \gamma,~ \forall i=1,...,u$. So,
\begin{equation*} \label{eq1}\scriptsize
\begin{split}
\sum_{j}\hat{T}_{uu_{ij}}^n &= \sum_j \sum_k \hat{T}_{uu_{ik}}^{(n-1)} \hat{T}_{uu_{kj}}\\
                           &= \sum_k \hat{T}_{uu_{ik}}^{(n-1)} \sum_j \hat{T}_{uu_{ik}}\\
                              &\leq \sum_k  \hat{T}_{uu_{ik}}^{(n-1)} \gamma \\
                              &\leq \gamma^n
\end{split}
\end{equation*}

Therefore, the sum of each row in $\hat{T}_{uu_{ij}}^n$ converges to zero, which indicates $\hat{T}_{uu_{ij}}^n Y^0 \leftarrow 0$.\\
\noindent{\bf Selection of $\sigma$.} Assuming a spatial representation of data points, we construct a minimum spanning tree using Kruskal's algorithm \cite{Kruskal1956} with distance between two nodes measured by Euclidean distance. Initially, no nodes are connected. We keep on adding edges in increasing order of distance. We choose the distance (say, $d_f$) of the first edge which connects two components with different labeled points in them. We consider $d_f$ as a heuristic to the minimum distance between two classes, and arbitrarily set $\sigma=d_0/3$, following $3\sigma$ rule of normal distribution \cite{Pukelsheim94}.

\begin{table*}[!ht]
\centering
\caption{Manually curated lists of words collected from analyzing the reference contexts. The lists are further expanded using the Wordnet:Synonym with different lexical variations. 
Note that while searching the occurrence of these words in reference contexts, we use different lexical variations of the words instead of exact matching.}\label{manual_list}

\scalebox{0.6}{
 \begin{tabular}{| l| l|}
 \hline
 {\bf \texttt{Rel}} & pivotal, comparable, innovative, relevant, relevantly, inspiring, related, relatedly, similar, similarly,              		    applicable, appropriate,  \\
                     &  pertinent, influential, influenced,  original, originally, useful, suggested, interesting,  inspired, likewise\\\hline

                    & recent, recently,  latest, later, late, latest, up-to-date, continuing, continued, upcoming, expected,  update, renewed,  	 extended\\
{\bf \texttt{Rec}}  & subsequent, subsequently,  initial, initially,  sudden, current, currently, future, unexpected, previous, previously, old, \\
		    & ongoing, imminent, anticipated, unprecedented, proposed, startling, preliminary, ensuing, repeated, reported, new,  earlier, \\
		    & earliest,  early, existing, further,  revised, improved  \\\hline
		
{\bf \texttt{Ext}}  & greatly,  awfully, drastically, intensely, acutely, almighty, exceptionally, excessively, exceedingly, tremendously, 				    importantly\\
                    & significantly, notably, outstandingly\\\hline
{\bf \texttt{Comp}} & easy, easier, easiest, vague, vaguer, vaguest, weak, weaker, weakest, strong, stronger, strongest, bogus, unclear \\\hline

 \end{tabular}}
 \vspace{-5mm}

\end{table*}

 \subsection{Features for Learning Model}\label{feature}
We use a wide range of features that suitably represent a paper-reference pair ($P_i,R_{ij}$), indicating $P_i$ refers to $P_j$ through reference $R_{ij}$. These features can be grouped into six general classes.
\vspace{-3mm}
\subsubsection{Context-based Features (CF)}
The ``reference context'' of $R_{ij}$ in $P_i$ is defined by three-sentence window  (sentence where $R_{ij}$ occurs and its immediate previous and next sentences). For multiple occurrences, we calculate its average score.
We refer to ``reference sentence'' to indicate the sentence where $R_{ij}$ appears.\\
(i) {\em \underline{CF:Alone}.} It indicates whether $R_{ij}$ is mentioned alone in the reference context or together with other references. \\
(ii) {\em \underline{CF:First}.} When $R_{ij}$ is grouped with others, this feature indicates whether it is mentioned first (e.g., ``[2]'' is first in ``[2,4,6]'').

Next four features are based on the occurrence of words in the corresponding lists created manually (see Table \ref{manual_list}) to understand different aspects. \\
(iii) {\em \underline{CF:Relevant}.} It indicates whether $R_{ij}$ is explicitly mentioned as relevant in the reference context (\texttt{Rel} in Table \ref{manual_list}).\\
(iv) {\em \underline{CF:Recent}.} It tells whether the reference context indicates that $R_{ij}$ is new (\texttt{Rec} in Table \ref{manual_list}).\\
(v) {\em \underline{CF:Extreme}.} It implies that $R_{ij}$ is extreme in some way (\texttt{Ext} in Table \ref{manual_list}).\\
(vi) {\em \underline{CF:Comp}.} It indicates whether the reference context makes some kind of comparison with $R_{ij}$ (\texttt{Comp} in Table \ref{manual_list}).

Note we do not consider any sentiment-based features as suggested by \cite{ASI:ASI23179}.
\vspace{-3mm}
\subsubsection{Similarity-based Features (SF)}\label{sf}
It is natural that the high degree of semantic similarity between the contents of $P_i$ and $P_j$ indicates the influence of $P_j$ in $P_i$. We assume that although the full text of $P_i$ is given, we do not have access to the full text of $P_j$ (may be due to the subscription charge or the unavailability of the older papers). Therefore, we consider only the title of $P_j$ as a proxy of its full text.  Then we calculate the cosine-similarity\footnote{We use the vector space based model \cite{Turney:2010} after stemming the words using Porter stammer \cite{Porter:1997}.} between the title ({\em T}) of $P_j$ and (i) {\em \underline{SF:TTitle.}} the title, (ii) {\em \underline{SF:TAbs}.} the abstract, {\em \underline{SF:TIntro}.} the introduction, (iv) {\em \underline{SF:TConcl}.} the conclusion, and (v) {\em \underline{SF:TRest}.} the rest of the sections (sections other than abstract, introduction and conclusion) of $P_i$.

We further assume that the ``reference context'' ({\em RC}) of $P_j$ in $P_i$ might provide an alternate way of summarizing the usage of the reference. Therefore, we take the same similarity based approach mentioned above, but replace the title of $P_j$ with its {\em RC} and obtain five more features: (vi) {\em \underline{SF:RCTitle}}, (vii) {\em \underline{SF:RCAbs}}, (viii) {\em \underline{SF:RCIntro}}, (ix) {\em \underline{SF:RCConcl}} and (x) {\em \underline{SF:RCRest}}. 
If a reference appears multiple times in a citing paper, we consider the aggregation of all $RC$s together.

\subsubsection{Frequency-based Feature (FF)}
The underlying assumption of these features is that a reference which occurs more frequently in a citing paper is more influential than a single occurrence \cite{Singh:2015}. We count the frequency of $R_{ij}$ in (i) {\em \underline{FF:Whole}.} the entire content, (ii) {\em \underline{FF:Intro}.} the introduction, (iii) {\em \underline{FF:Rel}.} the related work, (iv) {\em \underline{FF:Rest}.} the rest of the sections (as mentioned in Section \ref{sf}) of $P_i$. We also introduce (v) {\em \underline{FF:Sec}.} to measure the fraction of different sections of $P_i$ where $R_{ij}$ occurs (assuming that appearance of $R_{ij}$ in different sections is more influential). These features are further normalized using the number of sentences in $P_i$ in order to avoid unnecessary bias on the size of the paper.

\subsubsection{Position-based Features (PF)}
Position of a reference in a paper might be a predictive clue to measure the influence \cite{ASI:ASI23179}. Intuitively, the earlier the reference appears in the paper, the more important it seems to us. For the first two features, we divide the entire paper into  two parts equally based on the sentence count and then see whether $R_{ij}$  appears (i) {\em \underline{PF:Begin}.} in the beginning or (ii) {\em \underline{PF:End}.} in the end of $P_i$. Importantly, if $R_{ij}$ appears multiple times in $P_i$, we consider the fraction of times it occurs in each part.

For the other two features, we take the entire paper, consider sentences as atomic units, and  measure position of the sentences where $R_{ij}$ appears, including (iii) {\em \underline{PF:Mean}.} mean position of appearance, (iv) {\em \underline{PF:Std}.} standard deviation of different appearances. These features are normalized by the total length (number of sentences) of $P_i$.
, thus ranging from $0$ (indicating beginning of $P_i$) to $1$ (indicating the end of $P_i$).

\subsubsection{Linguistic Features (LF)}
The linguistic evidences around the context of $R_{ij}$  sometimes provide clues to understand the intrinsic influence of $P_{j}$ on $P_i$. Here we consider word level and structural features.\\
(i) {\em \underline{LF:NGram}.} Different levels of $n$-grams ($1$-grams, $2$-grams and $3$-grams) are extracted from the reference context to see the effect of different word combination \cite{Athar:2012}.\\
(ii) {\em \underline{LF:POS}.} Part-of-speech (POS) tags of the words in the reference sentence are used as features \cite{JochimS12}.\\
(iii) {\em \underline{LF:Tense}.}  The main verb of the reference sentence is used as a feature \cite{Teufel:2006}. \\
(iv) {\em \underline{LF:Modal}.} The presence of modal verbs (e.g., ``can'', ``may'') often indicates the strength of the claims. Hence, we check  the presence of the modal verbs in the reference sentence.\\
(v) {\em \underline{LF:MainV}.} We use the main-verb of the reference sentence as a direct feature in the model.\\
(vi) {\em \underline{LF:hasBut}.} We check the presence of  conjunction ``but'', which is another clue to show less confidence on the cited paper.\\
(vii) {\em \underline{LF:DepRel}.} Following \cite{Athar:2012} we use all the dependencies present in the reference context, as given by the dependency parser \cite{marneffe06generating}.\\
(viii) {\em \underline{LF:POSP}.} \cite{Cailing} use seven regular expression patterns of POS tags to capture syntactic information; then seven boolean features mark the presence of these patterns. We also utilize the same regular expressions as shown below
\footnote{The meaning of each POS tag can be found in \url{http://nlp.stanford.edu/software/tagger.shtml}\cite{Toutanova:2000}.} 
with the examples (the empty parenthesis in each example indicates the presence of a reference token $R_{ij}$ in the corresponding sentence; while few examples are complete sentences, few are not):
		    \begin{itemize}\setlength\itemsep{0.1em} \scriptsize
 				 \item ``.*\textbackslash\textbackslash(\textbackslash\textbackslash) VV[DPZN].*'': {\it Chen \underline{()
				  showed} that cohesion is held in the vast majority of cases for English-French.}
  				\item ``.*(VHP\textbar VHZ) VV.*'': {\it while Cherry and Lin () \underline{have shown} it to be a strong
				  feature for word alignment...}
  				\item ``.*VH(D\textbar G\textbar N\textbar P\textbar Z) (RB )*VBN.*'': {\it Inducing features for taggers by
				clustering \underline{has been} tried by several researchers ().}
  				\item ``.*MD (RB )*VB(RB )* VVN.*'': {\it For example, the likelihood of those generative procedures
\underline{can be accumulated} to get the likelihood of the phrase pair ().}
  				\item ``[∧ IW.]*VB(D\textbar P\textbar Z) (RB )*VV[ND].*'': {\it Our experimental set-up \underline{is
modeled} after the human evaluation presented in ().}
  				\item ``(RB )*PP (RB )*V.*'': {\it \underline{We use} CRF () to perform this
tagging.}
  				\item ``.*VVG (NP )*(CC )*(NP ).*'': {\it \underline{Following ()}, we provide the annotators with only
short sentences: those with source sentences between 10 and 25 tokens long.}
  			\end{itemize}

These are all considered as Boolean features.  			
For each feature, we take all the possible evidences from all paper-reference pairs and prepare a vector. Then for each pair, we check the presence (absence) of tokens for the corresponding feature and mark the vector accordingly (which in turn produces a set of Boolean features).

\subsubsection{Miscellaneous Features (MS)}
This group provides other factors to explain why is a paper being cited.
(i) {\em \underline{MS:GCount}.} To answer whether a highly-cited paper has more academic influence on the citing paper than the one which is less cited, we measure the number of other papers (except $P_i$) citing $P_j$. \\
(ii) {\em \underline{MS:SelfC}.} To see the effect of self-citation, we check whether at least one author is common in both $P_i$ and $P_j$.\\
(iii) {\em \underline{MG:Time}.} The fact that older papers are rarely cited, may not stipulate that these are less influential. Therefore, we measure the difference of the publication years of $P_i$ and $P_j$. \\
(iv) {\em \underline{MG:CoCite}.} It measures the co-citation counts of $P_i$ and $P_j$ defined by $\frac{|R_i \cap R_j|}{|R_i \cup R_j|}$, which in turn answers the significance of reference-based similarity driving the academic influence \cite{small1973cocitation}.

Following  \cite{Witten:2005}, we further make one step normalization and divide each feature by its maximum value in all the entires.

\section{Dataset and Annotation}\label{data}
We use the AAN dataset \cite{DragomirR} which is an assemblage of papers
included in ACL related venues. The texts are preprocessed where sentences, paragraphs and sections are properly separated using different markers.  The filtered dataset contains
12,843 papers (on average 6.21 references per paper) and 11,092 unique authors.

Next we use {\em Parscit}~\cite{councill2008parscit} to identify the reference contexts from the dataset and then extract the section headings from all the papers. Then each section heading is mapped into one of the following broad categories using the method proposed by \cite{LiakataSDBR12}: Abstract, Introduction, Related Work, Conclusion and Rest.

\noindent\textbf{Dataset Labeling.} The hardest challenge in this task is that there is no publicly available dataset where references are annotated with the intensity value. Therefore, we constructed our own annotated dataset in two different ways. (i)~{\em Expert Annotation}: we requested members of our research group\footnote{All were researchers with the age between 25-45 working on document summarization, sentiment analysis, and text mining in NLP.} to participate in this survey. To facilitate the labeling process, we designed a portal where all the papers present in our dataset are enlisted in a drop-down menu. Upon selecting a paper, its corresponding references were shown with five possible intensity values.
The citing and cited papers are also linked to the original texts so that the annotators can read  the original papers.
A total of $20$ researchers participated and they were asked to label as many paper-reference pairs as they could based on the definitions of the intensity provided in Section \ref{intensity}. The annotation process went on for one month. Out of total 1640 pairs annotated, $1270$ pairs were taken such that each pair was annotated by at least two annotators, and the final intensity value of the pair was considered to be the average of the scores. The Pearson correlation and Kendell's $\tau$ among the annotators are $0.787$ and $0.712$ respectively.  (ii) {\em Author Annotation}: we believe that the authors of a paper are the best experts to judge the intensity of references present in the paper. With this intension, we launched a survey where we requested the authors whose papers are present in our dataset with significant numbers. We designed a web portal in similar fashion mentioned earlier; but  each author was only shown her own papers in the drop-down menu. Out of $35$ requests, $22$ authors responded and total $196$ pairs are annotated. This time we made sure that each paper-reference pair was annotated by only one author. The percentages of labels in the overall annotated dataset are as follows: 1: 9\%, 2: 74\%, 3: 9\%, 4: 3\%, 5: 4\%.

\begin{figure}[!h]
\centering
\scalebox{0.18}{
\includegraphics{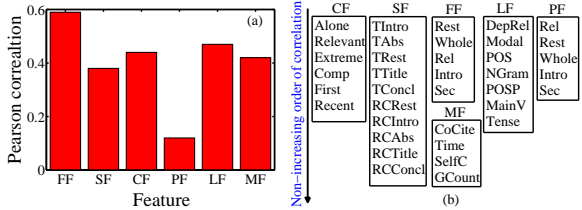}}
\caption{Pearson correlation coefficient between the features and the gold-standard annotations. (a) Group-wise average correlation, and (b) ranking of features in each group based on the correlation.}\label{corr_feature}
\vspace{-5mm}
\end{figure}
\vspace{-3mm}
\section{Experimental Results}\label{sec:result}
In this section, we start with analyzing the importance of the feature sets in predicting the reference intensity, followed by the detailed results.

\noindent{\bf Feature Analysis.} In order to determine which features highly determine the gold-standard labeling, 
we measure the Pearson correlation between various features and the ground-truth labels. Figure \ref{corr_feature}(a) shows the average correlation for each feature group, and in each group the rank of features based on the correlation is shown in Figure \ref{corr_feature}(b). Frequency-based features ({\em FF})  turn out to be the best, among which {\em FF:Rest} is mostly correlated. This set of features is convenient and can be easily computed. Both {\em CF} and {\em LF} seem to be equally important. However, $PF$ tends to be less important in this task.

\begin{table}[!h]
\caption{Performance of the competing models. The features are added greedily into the \gralap~ model.}\label{result}
 \subfloat[Baselines]{
 \scalebox{0.52}{
 \begin{tabular}{r |r |r |r}
 \hline
 Model  &   {\em RMSE} & $\rho$ & $R^2$ \\\hline
 \texttt{Uniform} & 2.09 & -0.05 & 3.21 \\
 \texttt{SVR+W}     & 1.95 & 0.54 &  1.34 \\
 \rowcolor[HTML]{96FFFB} 
 \texttt{SVR+O}     & 1.92 & 0.56 &  1.29 \\
 \texttt{C4.5SSL} & 1.99 & 0.46 & 2.46\\
 \texttt{GLM} & 1.98 & 0.52 & 1.35\\ \hline
 \end{tabular}}}
\quad
\subfloat[Our model]{
 \scalebox{0.55}{
 \begin{tabular}{l|r |r |r |r}
 \hline
 No. &Model  &   {\em RMSE} & $\rho$ & $R^2$ \\\hline
 (1) & \gralap + {\tt FF} & 1.10   & 0.79    & 1.05      \\
 (2) & (1) + {\tt LF} & 0.98  & 0.84     &  0.95   \\
 (3) & (2) + {\tt CF} & 0.90 &  0.87   & 0.87    \\
 (4)  & (3) + {\tt MF}  & 0.95 & 0.89 & 0.84  \\
 (5)  & (4) + {\tt SF}  & 0.92 & 0.90 & 0.82  \\
 \rowcolor[HTML]{96FFFB} 
 (6)  & (5) + {\tt PF}  & 0.91 & 0.90 & 0.80  \\\hline 
 \end{tabular}}}
 \vspace{-5mm}
\end{table}

\noindent{\bf Results of Predictive Models.}
For the purpose of evaluation, we report the average results after 10-fold cross-validation. Here we consider five baselines
to compare with \gralap: (i) {\tt Uniform}: assign $3$ to all the references assuming equal intensity, (ii) {\tt SVR+W}: recently proposed Support Vector Regression (SVR) with the feature set mentioned in  \cite{WanL14}, (iii) {\tt SVR+O}: SVR model with our feature set,  (iv) {\tt C4.5SSL}: C4.5 semi-supervised algorithm with our feature set \cite{Quinlan:1993:CPM:152181}, and (v) {\tt GLM}: the traditional graph-based LP model with our feature set \cite{Zhu03semi}. Three metrics are used to compare the results of the competing models with the annotated labels: {\em Root Mean Square Error} ({\em RMSE}), {\em Pearson's correlation coefficient} ($\rho$), and {\em coefficient of determination} ($R^2$)\footnote{The less ({\em resp.} more) the value of $RMSE$ and $R^2$ ({\em resp.} $\rho$), the better the performance of the models.}.

Table \ref{result} shows the performance of the competing models. We incrementally include each feature set into \gralap~greedily on the basis of ranking shown in Figure \ref{corr_feature}(a). We observe that \gralap~ with only {\tt FF} outperforms \texttt{SVR+O} with 41\% improvement of $\rho$. As expected, the inclusion of {\tt PF} into the model improves the model marginally. However, the overall performance of \gralap~ is significantly higher than any of the baselines ($p<0.01$).

\section{Applications of Reference Intensity}\label{appli}

In this section, we provide four different applications to show the use of measuring the intensity of references. To this end, we consider all the labeled entries for training and run \gralap~ to predict the intensity of rest of the paper-reference pairs.

\subsection{Discovering Influential Articles}\label{inf}
Influential papers in a particular area are often discovered by considering {\em equal weights} to all the citations of a paper. We anticipate that considering the reference intensity would perhaps return more meaningful results. To show this,
Here we use the following measures individually to compute the influence of a paper: (i) \texttt{RawCite:} total number of citations per paper, (ii) \texttt{RawPR:} we construct a citation network (nodes: papers, links: citations), and measure PageRank \cite{Pageetal98} of each node $n$: $PR(n)=\frac{1-q}{N}+q\sum_{m\in M(n)} \frac{PR(m)}{|L(m)|}$; where, $q$, the damping factor, is set to 0.85, $N$ is the total number of nodes, $M(n)$ is the set of nodes that have edges to $n$, and $L(m)$ is the set of nodes that $m$ has an edge to, (iii) \texttt{InfCite:} the weighted version of \texttt{RawCite}, measured by the sum of intensities of all citations of a paper, (iv) \texttt{InfPR:} the weighted  version of \texttt{RawPR}:  $PR(n)=\frac{1-q}{N}+q\sum_{m\in M(n)} \frac{Inf(m\rightarrow n)PR(m)}{\sum_{a\in L(m) Inf(m \rightarrow a)}}$, where $Inf$ indicates the influence of a reference. We rank all the articles based on these four measures separately. Table \ref{corr}(a) shows the Spearman's rank correlation between pair-wise measures. As expected, (i) and (ii) have high correlation (same for (iii) and (iv)), whereas across two types of measures the correlation is less. Further, in order to know which measure is more relevant, we conduct a subjective study where we select top ten papers from each measure and invite the experts (not authors) who annotated the dataset,  to make a binary decision whether a recommended paper is relevant.
\footnote{We choose papers from the area of ``sentiment analysis'' on which experts agree on evaluating the papers.}. 
The average pair-wise inter-annotator's  agreement (based on Cohen's kappa \cite{cohen1960}) is $0.71$. Table \ref{corr}(b) presents that out of $10$ recommendations of \texttt{InfPR}, $7$ ($5$) papers are marked as influential by majority (all) of the annotators, which is followed by \texttt{InfCite}. 
These results indeed show the utility of measuring reference intensity for discovering influential papers.
Top three papers based on \texttt{InfPR} from the entire dataset are shown in Table \ref{list}.

\vspace{-2mm}
\begin{table}[!h]
\caption{(a) Spearman's rank correlation among influence measures and (b) expert evaluation of the ranked results (for top $10$ recommendations).}\label{corr}
 \subfloat[]{
 \scalebox{0.49}{
 \begin{tabular}{|c|c|c|c|c|}
 \hline
                  & \texttt{RowCite} & \texttt{RowPR} & \texttt{InfCite} & \texttt{InfPR} \\\hline
\texttt{RowCite} & 1 & 0.82 & 0.61 & 0.54 \\\hline
\texttt{RowPR} & 0.82 & 1 & 0.52 & 0.63 \\\hline
\texttt{InfCite} & 0.61 & 0.52 & 1 & 0.84\\\hline
\texttt{InfPR} &  0.54 & 0.63 &  0.84 & 1\\ \hline
 \end{tabular}}}
\quad
\subfloat[]{
 \scalebox{0.49}{
 \begin{tabular}{|c|c|c|}
 \hline
           Metric       & All & Majority \\\hline
\texttt{RowCite} & 2 & 5 \\
\texttt{RowPR} & 2 & 4  \\
\texttt{InfCite} & 4 & 5 \\
\rowcolor[HTML]{96FFFB} 
\texttt{InfPR} &  5 & 7 \\\hline
 \end{tabular}}}
 \vspace{-10mm}
\end{table}

\if{0}
\begin{table}[!h]
 \caption{Expert evaluation of the ranked results (top $10$) based on different influence measures.}\label{exp}
 \scalebox{0.8}{
 \begin{tabular}{|c|c|c|c|c|}
 \hline
Annotators  & \texttt{RawCite} & \texttt{RawPR} & \texttt{InfCite} & \texttt{InfPR} \\\hline
 All       &  2            &       2          &  4            & 5  \\\hline
Majority   & 5             &     4            5               & 7   \\\hline
 \end{tabular}}
\end{table}
\fi

\begin{table*}[!th]
\centering
 \caption{Top three papers and authors based on \texttt{InfPR} and {\tt Hif-index} respectively.}\label{list}
 \scalebox{0.7}{
 \begin{tabular}{c|l|l}
 \hline
No & \multicolumn{1}{c|}{Paper}  & \multicolumn{1}{c}{Author} \\\hline
1. & Lexical semantic techniques for corpus analysis \cite{Pustejovsky:1993}& Mark Johnson\\
2. & An unsupervised method for detecting grammatical errors \cite{Chodorow:} & Christopher D. Manning\\
3. & A maximum entropy approach to natural language processing \cite{Berger:1996} & Dan Klein\\\hline

\end{tabular}}
\vspace{-5mm}
\end{table*}

\subsection{Identifying Influential Authors}
H-index, a measure of impact/influence of an author, considers each citation with equal weight \cite{hirsch2005index}. Here we incorporate the notion of reference intensity into it and define  {\tt hif-index}.

\begin{mydef}\small
 An author $A$ with a set of papers $P(A)$ has an {\tt hif-index} equals to $h$, if $h$ is the largest value such that $|\{p \in P(A)|Inf(p)\geq h\}|\geq h$; where $Inf(p)$ is the sum of intensities of all citations of $p$.
\end{mydef}

\begin{figure}[!h]
\centering
\scalebox{0.18}{
\includegraphics{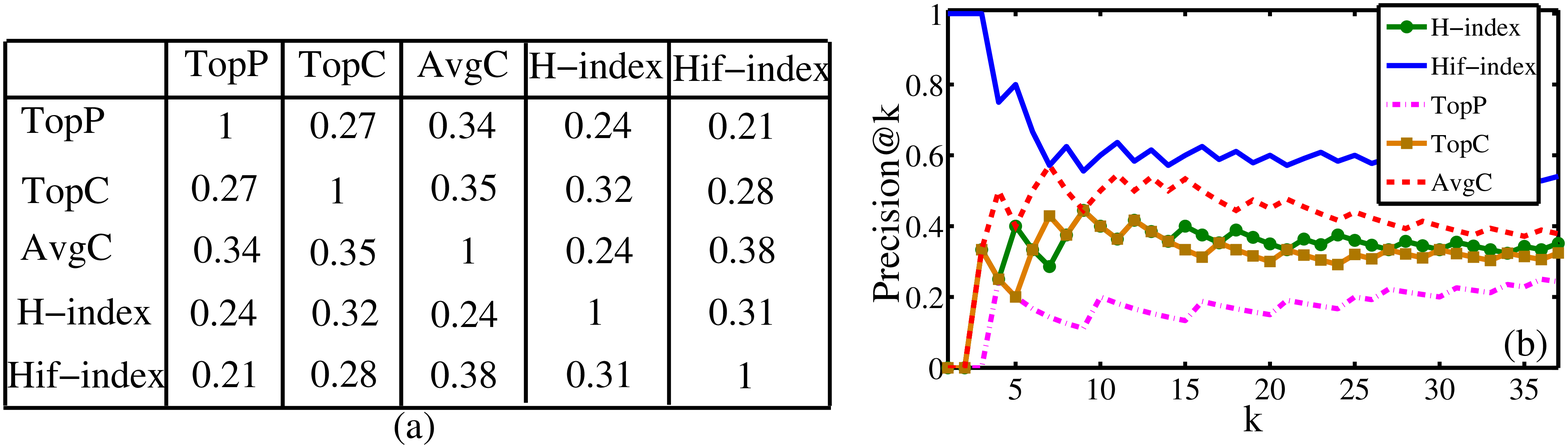}}
\caption{(a) Sprearman's rank correlation among pair-wise ranks, and (b) the performance of all the measures.}\label{rank}
\vspace{-6mm}
\end{figure}

We consider $37$ ACL fellows as the list of gold-standard influential authors. For comparative evaluation, we consider the total number of papers (\texttt{TotP}), total number of citations (\texttt{TotC}) and average citations per paper (\texttt{AvgC}) as three competing measures along with {\tt h-index} and {\tt hif-index}. We  arrange all the authors in our dataset in decreasing order of each measure. Figure \ref{rank}(a) shows the Spearman's rank correlation among the common elements across pair-wise rankings. Figure \ref{rank}(b) shows the $Precision@k$ for five competing measures at identifying ACL fellows. We observe that {\tt hif-index} performs significantly well with an overall precision of $0.54$, followed by \texttt{AvgC} ($0.37$), h-index ($0.35$), \texttt{TotC} ($0.32$) and \texttt{TotP} ($0.34$). This result is an encouraging evidence that the reference-intensity could improve the identification of the influential authors. 
Top three authors based on {\tt hif-index} are shown in Table \ref{list}.

\vspace{-3mm}

\subsection{Effect on Recommendation System}
Here we show the effectiveness of reference-intensity by applying it to a real paper recommendation system. To this end, we consider \texttt{FeRoSA}\footnote{\url{www.ferosa.org}} \cite{ferosa}, a new (probably the first) framework of faceted recommendation for scientific articles, where given a query it provides facet-wise recommendations with each facet representing the purpose of recommendation \cite{ferosa}.  The methodology is based
on random walk with restarts (RWR) initiated from a query paper. The model is built on AAN dataset and considers both the citation links and the content information to produce the most relevant results. Instead of using the unweighted citation network, here we use the weighted network with each edge labeled by the intensity score. The final recommendation of \texttt{FeRoSA} is obtained by performing RWR with the transition probability proportional to the edge-weight (we call it \texttt{Inf-FeRoSA}).  We observe that \texttt{Inf-FeRoSA} achieves an average precision of $0.81$ at top 10 recommendations, which is 14\% higher then {\tt FeRoSA} while considering the flat version and 12.34\% higher than \texttt{FeRoSA} while considering the faceted version.

\begin{figure}[!h]
\centering
\includegraphics[width=\columnwidth]{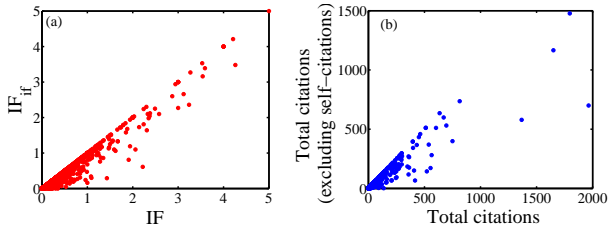}
\caption{Correlation between (a) $IF$ and $IF_{if}$ and (b) number of citations before and after removing self-journal citations.}\label{scatter}
\vspace{-3mm}
\end{figure}

\subsection{Detecting Citation Stacking}


Recently, Thomson Reuters began screening for journals that exchange large number of anomalous citations with other journals in a cartel-like arrangement, often known as ``citation stacking'' \cite{cartel1,cartel2}. This sort of citation stacking is much more pernicious and difficult to detect. We anticipate that this behavior can be detected by the reference intensity. Since the AAN dataset does not have journal information, we use DBLP dataset \cite{Singh:2015} where the complete metadata information (along with reference contexts and abstract) is available, except the full content of the paper (559,338 papers and 681 journals; more details in \cite{Chakraborty:2014}). From this dataset, we extract all the features mentioned in Section \ref{feature} except the ones that require full text, and run our model using the existing annotated dataset as training instances. We measure the traditional impact factor ($IF$) of the journals and impact factor after considering the reference intensity ($IF_{if}$). Figure \ref{scatter}(a) shows that there are few journals whose $IF_{if}$ significantly deviates (3$\sigma$ from the mean) from $IF$; out of the suspected journals 70\% suffer from the effect of self-journal citations as well
(shown in Figure \ref{scatter}(b)), example including {\em Expert Systems with Applications} (current $IF$ of $2.53$). One of the future work directions would be to predict such journals as early as possible after their first appearance.

\section{Related Work}

Although the citation count based metrics are widely accepted \cite{Eugene,Hirsch:2010}, the belief that mere counting of citations is dubious has also been a subject of study \cite{Chubin}. \cite{Garfield64cancitation} was the first who explained the reasons of citing a paper. 
\cite{pham} introduced a method for the rapid development of complex rule bases for classifying text segments. 
\cite{Cailing} focused on a less manual approach by learning domain-insensitive features from textual, physical, and syntactic aspects
To address concerns about h-index, different alternative measures are proposed \cite{Waltman:2012}.
However they too could benefit from filtering or weighting references with a model of influence. 
Several research have been proposed to weight citations based on factors such as the prestige of the citing journal \cite{Ding11,RePEc:bla}, prestige of an author \cite{Balaban2012}, frequency of citations in citing papers \cite{BIES:BIES201100067}. 
Recently, \cite{WanL14} proposed a SVR based approach to measure the intensity of citations. Our methodology differs from this approach in at lease four significant ways: (i) they used six very shallow level features; whereas we consider features from different dimensions, (ii) they labeled the dataset by the help of independent annotators; here we additionally ask the authors of the citing papers to identify the influential references which  is very realistic \cite{gil}; (iii) they adopted SVR for labeling, which does not perform well for small training instances; here we propose \gralap~, designed specifically for small training instances; (iv) four applications of reference intensity mentioned here are completely new and can trigger further to reassessing the existing bibliometrics.

\section{Conclusion}

We argued that the equal weight of all references might not be a good idea not only to gauge success of a research, but also to track follow-up
work or recommending research papers. The annotated dataset would have tremendous potential to be utilized for other research. Moreover, \gralap~ can be used for any semi-supervised learning problem. Each application mentioned here needs separate attention. In future, we shall look into more linguistic evidences to improve our model.

\end{document}